\documentclass[10pt,letterpaper]{article}

\usepackage{cogsci}

 \cogscifinalcopy 

\usepackage[
  style=apa,
  natbib=true,
  annotation=false,
]{biblatex}
\usepackage{float} 

\usepackage{algorithm2e}
\usepackage{bbm}
\usepackage{graphicx}
\usepackage{comment}
\usepackage{booktabs}
\usepackage{xcolor}

\RestyleAlgo{ruled}

\title{Evaluating the relationship between regularity and learnability in recursive numeral systems using Reinforcement Learning}

\author{
  {\large\bfseries Andrea Silvi\textsuperscript{1} \quad 
  Ponrawee Prasertsom\textsuperscript{2}
  \quad 
  Jennifer Culbertson\textsuperscript{2}} \\
  {\large\bfseries {Devdatt Dubhashi\textsuperscript{1}} \quad 
  {Moa Johansson\textsuperscript{1}} \quad 
 {Kenny Smith\textsuperscript{2}}}\vspace{0.2cm}
  \\
   \texttt{\{silvi, moa.johansson, dubhashi\}@chalmers.se}\\
   \texttt{\{ponrawee.prasertsom, jennifer.culbertson, kenny.smith\}@ed.ac.uk}\\
  {\normalsize\normalfont
      \textsuperscript{1}Chalmers University of Technology and Gothenburg University  \quad
      \textsuperscript{2}University of Edinburgh
  }
}

\begin{document}

\maketitle

\begin{abstract}
Human recursive numeral systems (i.e., counting systems such as English base-10 numerals), like many other grammatical systems, are highly regular.
Following prior work that relates cross-linguistic tendencies to biases in learning, we ask whether regular systems are common because regularity facilitates learning.
Adopting methods from the Reinforcement Learning literature, we confirm that highly regular human(-like) systems are easier to learn than unattested but possible irregular systems. This asymmetry emerges under the natural assumption that recursive numeral systems are designed for generalisation from limited data to represent all integers exactly.
We also find that the influence of regularity on learnability is absent for unnatural, highly irregular systems, whose learnability is influenced instead by signal length, suggesting that different pressures may influence learnability differently in different parts of the space of possible numeral systems.
Our results contribute to the body of work linking learnability to cross-linguistic prevalence.

\end{abstract}

\section{Introduction}
Linguistic systems are highly diverse, nevertheless they also exhibit similarities \citep{evans2009myth,croft2002typology}.
Research has shown that cross-linguistically prevalent features can be explained by a pressure for efficiency \citep{gibson_how_2019,kemp2018semantic}.
In this paper, we focus on recursive numeral systems, i.e., systems of numerals that can express exact quantities (e.g., English \textit{one, two, three, \ldots, ten, eleven, twelve, \ldots, twenty one, twenty two, twenty three, \ldots}).
\citet{prasertsom2026recursivenumeralsystemshighly} show that natural recursive numeral systems, as opposed to theoretically possible but unattested systems, are characterised by their efficiency; while such systems are maximally informative (i.e., any integer can be precisely conveyed), they exhibit a trade-off between processing complexity and regularity. Here, we focus on the latter, and explore its connection to learnability.

Previous research has connected learnability to cross-linguistic prevalence across a number of linguistic domains using artificial language learning experiments with humans, and computational models \citep[e.g.,][]{culbertson2012learning,steinert2020ease}. We investigate whether regularity in recursive numeral systems is driven by a pressure for learnability, using Reinforcement Learning (RL) as a tool to measure learnability. We ask whether computational agents can learn human-like, regular recursive numeral systems better than irregular ones.
We find that regular recursive numeral systems are indeed more learnable than unnatural irregular ones in an RL setting. Importantly this result depends on the assumption that ability to generalise from limited training data to express a larger range of numbers is an important aspect of learnability. 
Interestingly, this holds in particular for systems that are highly regular, like human recursive numeral systems; regularity has less of an effect on the learnability of highly irregular systems.

We begin by briefly reviewing three relevant strands of research on language learning, cross-linguistic tendencies, and RL, which provide a basis for our study.

\subsection{Efficiency of recursive numeral systems}
Recent research seeks to explain why linguistic systems are the way they are by appealing to the fact that they are optimised for efficient communication \citep{gibson_how_2019,kemp2018semantic}, i.e., they minimise complexity while maximising communcative accuracy.
A recent line of work has looked at numeral systems under this lens.
\citet{denic2024recursive,yang2025re}  argue that recursive numeral systems optimise the trade-off between lexicon size (the number of unique morphemes that are used in the numeral system) and average morphosyntactic complexity (average number of morphemes per numeral). 
Intuitively, a system with a large lexicon consisting of many unique monomorphemic \textit{numerals} for specific \textit{numbers} can get by with shorter numerals (e.g. in a system with unique numerals for numbers in the range 1--20, all numbers in that range can be expressed with a single monomorphemic form of morphosyntactic complexity 1); a system with a smaller lexicon (e.g. dedicated morphemes only for numbers 1--10) needs to use longer numerals to express larger numbers (e.g. 11--20 might be expressed by numerals consisting of ``ten plus one'', ``ten plus two" etc).

\begin{table}[ht]
\footnotesize
\centering
\begin{tabular}{crr}
\toprule
Number & Numeral (Mandarin) & Numeral (unnatural) \\
\midrule
18 & $10 + 8$ & $2 * 9$  \\
19 & $10 + 9$ & $10 + 9$  \\
20 & $2 * 10$ & $2 * 10$  \\
21 & $2 * 10 + 1$ & $2 * 9 + 3$ \\
22 & $2*10+2$ & $2 * 10 + 2$ \\
23 & $2*10+3$ & $2 * 9 + 5$ \\
24 & $2*10+4$ & $2 * 10 + 4$ \\
25 & $2*10+5$ & $2 * 9 + 7$ \\
\bottomrule
\end{tabular}
\caption{Numerals in Mandarin (first column) and in a hypothetical system (second column) that is matched for lexicon size and average morphosyntactic complexity for the number range $18-25$. Notice that Mandarin is consistently base-10, whereas the unnatural system has alternating bases (9, 10).}
\label{tab:ex_mandarin_possible}
\end{table}

While this trade-off has some explanatory power, \citet{prasertsom2026recursivenumeralsystemshighly} argue that it fails to take into account the regularity of numeral forms. 
Natural numeral systems tend to be highly regular (e.g. Mandarin in Table \ref{tab:ex_mandarin_possible}), but the measures used by \citet{denic2024recursive} assign the same level of efficiency to highly irregular, unnatural systems (see example hypothetical system in Table \ref{tab:ex_mandarin_possible}), as long as they have the same lexicon size and average length.

Drawing on the Minimum Description Length approach from \citet{brighton2003simplicity}, \citet{prasertsom2026recursivenumeralsystemshighly} propose that these systems are thus better thought of as being subject to competing pressures to maximise \textit{regularity} (how much form reuse there is) and minimise \textit{processing complexity} (how difficult it is to parse or generate numerals).
This approach is able to delineate natural recursive numeral systems from theoretically possible but unattested ones, with the natural ones exhibiting higher regularity and lower processing complexity.

\subsection{Simplicity, learnability and cross-linguistic prevalence}

\citet{prasertsom2026recursivenumeralsystemshighly} also suggest that \textit{learning} may be the mechanism through which recursive numeral systems are optimised for regularity: regular systems have short description lengths, which make them easier to learn, more stable in intergenerational transmission, and therefore more widespread. 
There is a wealth of experimental evidence showing that simpler systems (i.e. those than can be described in a shorter or more compressed way in a given description language) are easier for humans to learn (\citealp[e.g.][]{shepard1961learning, raviv2021,wang2025learning,moreton2017phonological}; see \citealp{CHATER200319} for a broad review of simplicity principles in cognition). 
Experiments have also shown that learners introduce regularity into irregular linguistic systems \citep[e.g.,][]{smith2010eliminating, keogh2024predictability}. 


Similar results also often arise in machine learning, where more human-like systems are more machine-learnable \citep{steinert2019learnability,steinert2020ease, Galke_2024, osmelak2026systematicityformsmeaningslanguages}.
This suggests that machine learning models may suitably approximate human learning in this respect.
We therefore hypothesise that human(-like), simpler, regular recursive numeral systems should also be easier for simulated agents (and thus plausibly also humans) to learn.
Next, we turn to how learnability can be measured in this domain.

\subsection{Measuring learnability via reinforcement learning}
RL has been used as a tool to study the emergence of communication \citep{foerster2016, LazaridouPB17, mordatch2018}, often modelling the evolution of language as a cooperative process entailing the maximisation of  future rewards.
In this area, a growing body of work has shown that the most communicatively efficient languages are also the ones that are easiest for deep RL agents to learn and converge on \citep{kageback2020eff, carlsson2021learning, carlsson2024, silvi2025le}.
In other words, one could use RL to study the learnability of different theoretical types of linguistic systems.

Specifically, the learnability of a system can be quantified by how rapidly a single agent learns to minimise errors given a fixed number of learning exposures or budget of interactions with the environment.
To measure this, theoretical work often adopts sample complexity (i.e., the number of samples required for convergence) to derive asymptotic bounds \citep{kakade2003sample}.
Similarly, empirical studies quantify efficiency by computing the Area Under the Curve (AUC) of  learning statistics \citep{taylor2009,agarwal2021}.
We adopt the latter approach to quantify the learnability of recursive numeral systems.





\section{Experiment 1: Learnability of recursive numerals with varying regularity}
First, we investigate the learnability of recursive numeral systems with different degrees of regularity.
We expect that regularity and learnability under RL should correlate.

\subsection{Methods}
Across all experiments, we use single-agent RL \citep{Sutton1998}, where an agent tries to maximise cumulative reward over iterations (epochs).
The goal is to learn to map each numeral (i.e., the form) to their number (i.e., the numerical quantity) in a given recursive numeral system.
At each iteration, a batch of numerals are sampled with replacement from a training distribution.
A single RL agent processes the numerals morpheme by morpheme and predicts numbers between 1--99,\footnote{For tractability and comparability with previous work, we only consider the numerals from 1  to 99.} and is rewarded depending on how close the predicted number is to the target number.

More formally, we characterise the task of learning recursive numerals as a contextual bandit problem with continuous action space \citep{agrawal1995continuum, lattimore2020bandit}.
Our agent is a single-layer LSTM network \citep{hochreiter1997long}.
Each morpheme in a numeral is represented by a learned size-5 embedding, and the resulting sequence of embeddings is processed one-by-one by the LSTM (hidden size of 10).
We use the last hidden state to predict a scalar mean $\mu{} \in [0.0, 1.0]$, which is interpreted as the mean of a Gaussian policy $\mathcal{N}(\mu{}, \sigma{})$.
We decrease the standard deviation $\sigma{}$ as the training proceeds to encourage exploration early while encouraging exploitation later.
The distribution is then used to sample a prediction, which is then multiplied by 100 to obtain a number between 1--99.
Our reward function $\mathcal{R}$
is bounded, exponentially-decaying, Gaussian-like:
\begin{equation}
    \mathcal{R}(\hat n, n)= \mathbbm{1}[0 < \text{int(}\hat n) < 100]\exp(-\alpha \lvert \hat n - n\rvert)
\end{equation}
where $\text{int(}\hat n)$ approximates $\hat n$ to its nearest integer, $\alpha=0.5$ determines the sharpness of the reward function. The first part of the reward function (the indicator function) ensures predictions outside the range $1-99$ are not rewarded.
The second part penalises predictions that are far away from the target, with the reward rapidly approaching and becoming effectively 0 if the distance is 10 or more.


 The RL agent is trained via the classical policy-gradient objective REINFORCE \citep{williams1992reinforce} over 30,000 epochs, with 5 batches of 32 numerals per epoch, sampled with replacement from the \textit{training distribution} $p(n)$.
The agent is tested on a sample of 99 numerals from the \textit{test distribution} $q(n)$.
The \textit{learnability} of a given recursive numeral system is the AUC of accuracy at test against epochs, i.e. the approximation of the integral under the curve obtained by plotting epoch against test accuracy, where higher AUC corresponds to more rapid learning.
We adopt this measure as it is more robust to training noise than other common ones such as final accuracy \citep{steinert2020ease}. 
This setup effectively frames the problem of learning numeral systems as a regression problem, 
rather than a classification one used in prior literature \citep{guo2019emergence, carlsson2021learning, silvi2025le}.
As previous classification approaches treat numbers as discrete orthogonal labels, the agent there would need to learn that some numbers are closer to each other than others via the reward function.
Under our regression approach, the agent does not have to learn this, as the output space already has the required ordering in place (i.e., it is already a number line).
For example, a classification task would require the agent to learn that the class for $15$ is close to the class for $16$ via a distance-based reward function, whereas in our regression task, there is no need for the agent to learn that 15 ($\mu$ = 0.15) is close to 16 ($\mu$ = 0.16), so the agent simply has to learn the form-meaning mappings.

\begin{figure}[t]
    \centering
    \includegraphics[width=0.98\columnwidth]{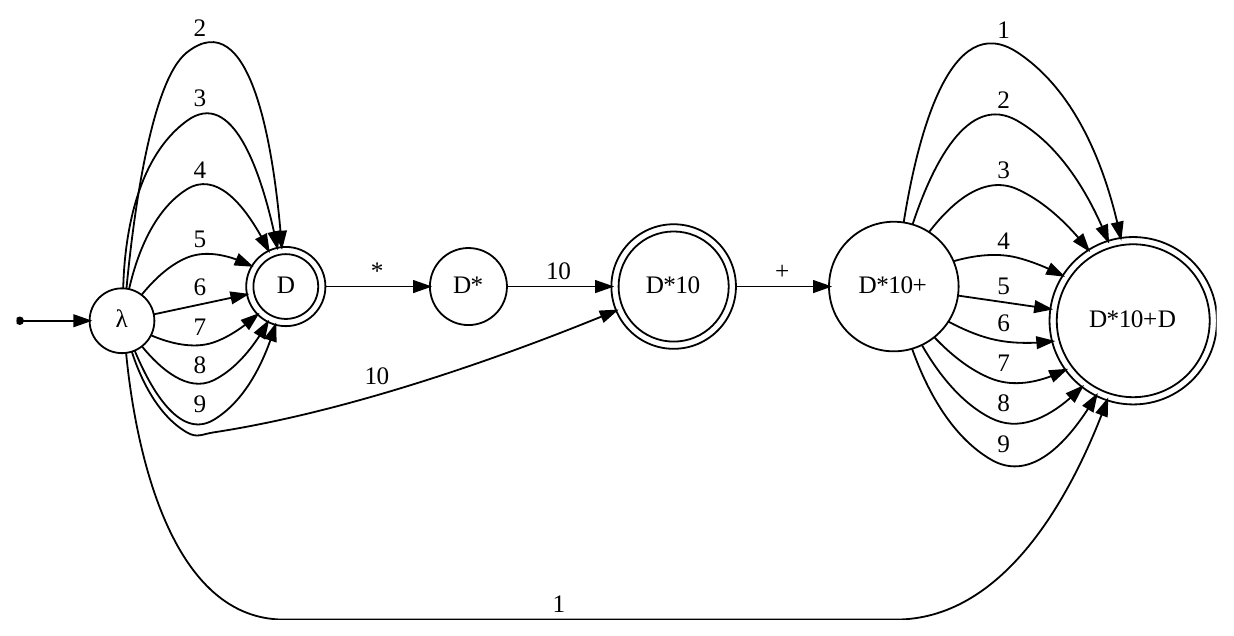}
    \caption{Inferred minimal partial DFA representing the Mandarin recursive numeral grammar.
    Transitions (arrows) represent output morphemes.
    States (circles) are labelled roughly according to the relevant grammar rules they capture.
    Double borders indicate accepting states.
    Generating the numeral for 20 ($2*10$) would involve going from the initial state ($\lambda{}$) to $D$ (emitting $2$), from $D$ to $D*$ (emitting $*$), and from $D*$ to $D*10$ (emitting $10$) and terminate the machine there.}
    \label{fig:mandarin}
\end{figure}

To investigate the effect of regularity on learnability, we follow the same formalisation of \textit{ir}regularity in \citet{prasertsom2026recursivenumeralsystemshighly}.
The measure rests on the intuition that regular systems can be represented withsmaller sets of grammar rules, which can be represented by a finite-state machine that generates all and only valid numeral strings.
Figure \ref{fig:mandarin} gives an example of an inferred  minimal partial DFA that generates all and only valid Mandarin numerals, a highly regular system.
The \textit{irregularity} is the size of the numeral grammar, measured by the number of bits required to encode the DFA: 
\begin{equation}\label{eq:lG}
    \underbrace{|Z|\bigl(2\cdot{}log_2|S| + log_2|\Sigma|\bigr)}_{\text{(A)}} + \underbrace{log_2|S|}_{\text{(B)}} + \underbrace{|S|}_{\text{(C)}}
\end{equation}
where (A) is the number of bits required to encode the transition table, where $Z$ is the set of transitions, $S$ the set of states and $\Sigma$ the set of symbols (morphemes).
(B) and (C) are the numbers of bits for encoding the identity of initial state and accepting states.
We use the same algorithm \citep{mihovFiniteStateTechniquesAutomata2019} and implementation (\texttt{automata-lib} in Python) for minimal DFA inference as \citeauthor{prasertsom2026recursivenumeralsystemshighly}.



\subsection{Experiment setup}
We quantify the relationship between learnability and regularity in the sample of recursive numeral systems investigated by  \citet{prasertsom2026recursivenumeralsystemshighly}.
This includes 1) 128 human recursive numeral systems from \citet{denic2024recursive}, 2) a sample of artificially generated systems under the lexicon size/average morphosyntatic complexity trade-off from \citet[data from][, henceforth, \textit{D\&S optimal systems}]{denic2024recursive}, 3) a sample of optimal artificial systems under the same trade-off with post-hoc constraints imposed to enforce human-likeness from \citet{yang2025re} (henceforth, \textit{Y\&R optimal systems}), and 4) a sample of 300 random baseline unattested systems that make use of the symbols (digits and multipliers) found in human systems. Note that all the artificial languages are either sampled or selected to have a lexicon size which is comparable to the ones attested in human systems (i.e., lexicon size of 5--13). To ensure robustness, we compute the learnability as described above 20 times for each system, taking the mean.

In this initial setup, our training and test distributions are both the power law distribution $p(n)\propto n^{-2}$, $q(n)\propto{} n^{-2}$, following previous work \citep{xu2020numeral, carlsson2021learning, denic2024recursive, silvi2025le}, which drew on the estimated communicative need of each number $n$ \citep{dehaene1992cross}.
Under this highly-skewed distribution, about 75\% of training and test samples will consist of numbers 1 and 2 alone, and about 90\% will consist of numbers in the range 1-6.



\begin{figure*}[ht]
    \centering
    \includegraphics[width=0.8\textwidth]{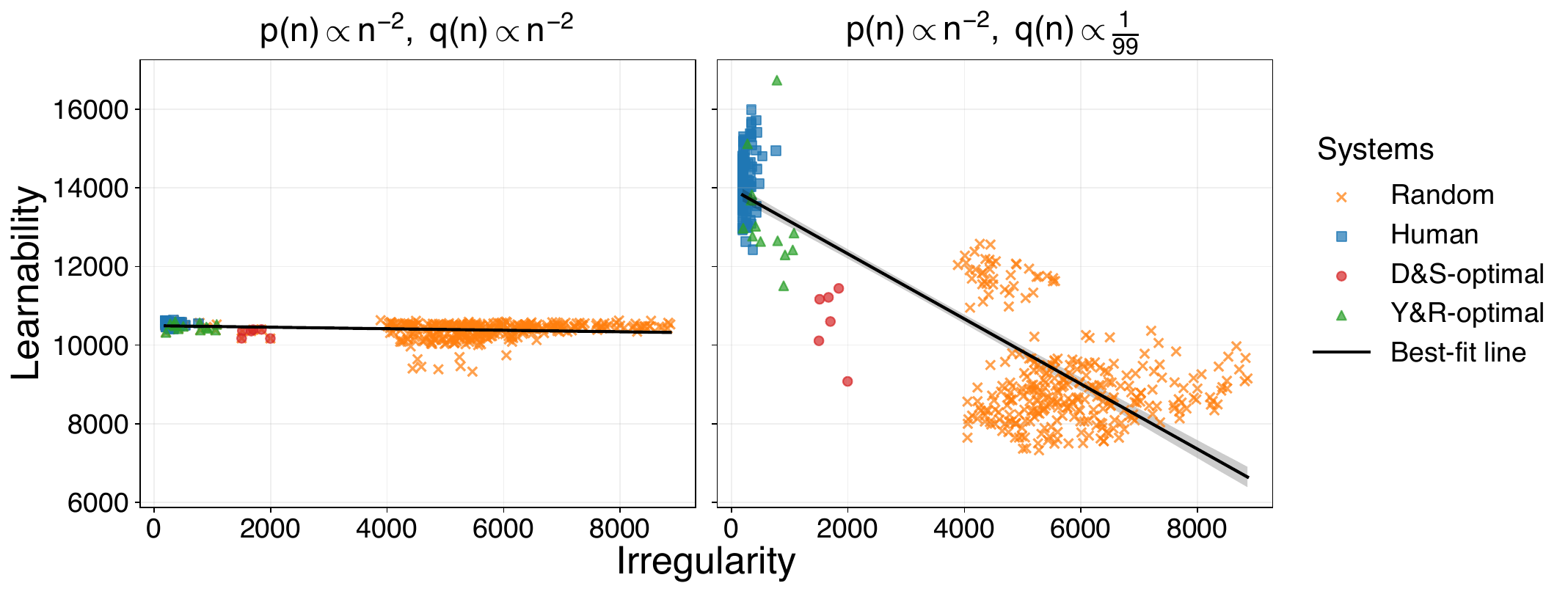}
    \caption{Irregularity and learnability for human (blue), D\&S optimal (red), Y\&R optimal (green), and random baseline (orange) recursive numeral systems under power law (left panel) and the uniform (right panel) test distributions $q(n)$. The training distribution $p(n)$ is always the power law distribution.}
    \label{fig:optimal_random_flat_and_not}
\end{figure*}

\subsection{Results and discussion}
Figure \ref{fig:optimal_random_flat_and_not} (left) plots irregularity against learnability, along with a least-square regression fit to this data.
Surprisingly, we found that there is little correlation (as indicated by the almost horizontal regression line of slope $\beta = -0.020$): there is no difference between regular, human(-like) systems and irregular systems in terms of learnability.

What could explain this result? We assumed that the primary distribution that shapes recursive numeral systems is the power law distribution derived from frequency of use.
This means that, at test, the agent trained on the system only has to perform well on the lower ranges of numerals in order for that system to score high on learnability.
However, as \citet{prasertsom2026recursivenumeralsystemshighly} point out, it is likely that recursive numeral systems are sensitive to the pressure to precisely communicate about \textit{any} integer, including very large numbers \citep[cf. also][]{frank2008piraha,pitt2022exact}.
This is at odds with using the power law distribution to evaluate learnability, under which larger numbers are virtually never sampled and which therefore contribute negligibly to measured learnability. 

\section{Experiment 2: Learnability as accuracy of generalisation}
We therefore separate out two relevant distributions that influence our measure of the learnability of recursive numeral systems.
One is the \textit{learning distribution}, which reflects the statistics of the input to our learners.
The other is the \textit{generalising distribution}, which we argue should reflects the need to precisely communicate all numbers.

In Experiment 2, we assume that the training distribution $p(n)$ is based on the frequency of use, and thus we use the power law distribution suggested in prior work. By contrast, $q(n)$ will be flat (an extreme case of a less-skewed distribution), reflecting the idea that numerals in the long tail also require high precision.
We thus explore the relationship between regularity and learnability in terms of the capacity to generalise to uniform $q(n)$. 

\subsection{Experiment setup}
We adopt the same methods as the previous experiment, but use a uniform distribution over $1-99$, i.e., $p(n) \propto{} n^{-2}$ and $q(n) \propto{} \frac{1}{99}$.



\subsection{Results and discussion}
Figure \ref{fig:optimal_random_flat_and_not} (right) plots the learnability of the systems under a uniform $q(n)$.
As expected, highly regular human systems now have on average the highest learnability and there is a negative correlation between irregularity and learnability ($\beta=-0.828$). 
This is likely because the agents can learn subparts of numerals and then reuse them to more efficiently learn the rarer, larger numbers. 
In contrast, the sampled random systems are harder to learn, likely because they tend to jump between bases, reusing the same sub-forms less frequently and less systematically. 

\begin{figure*}[t]
    \centering
    \includegraphics[width=0.8\textwidth]{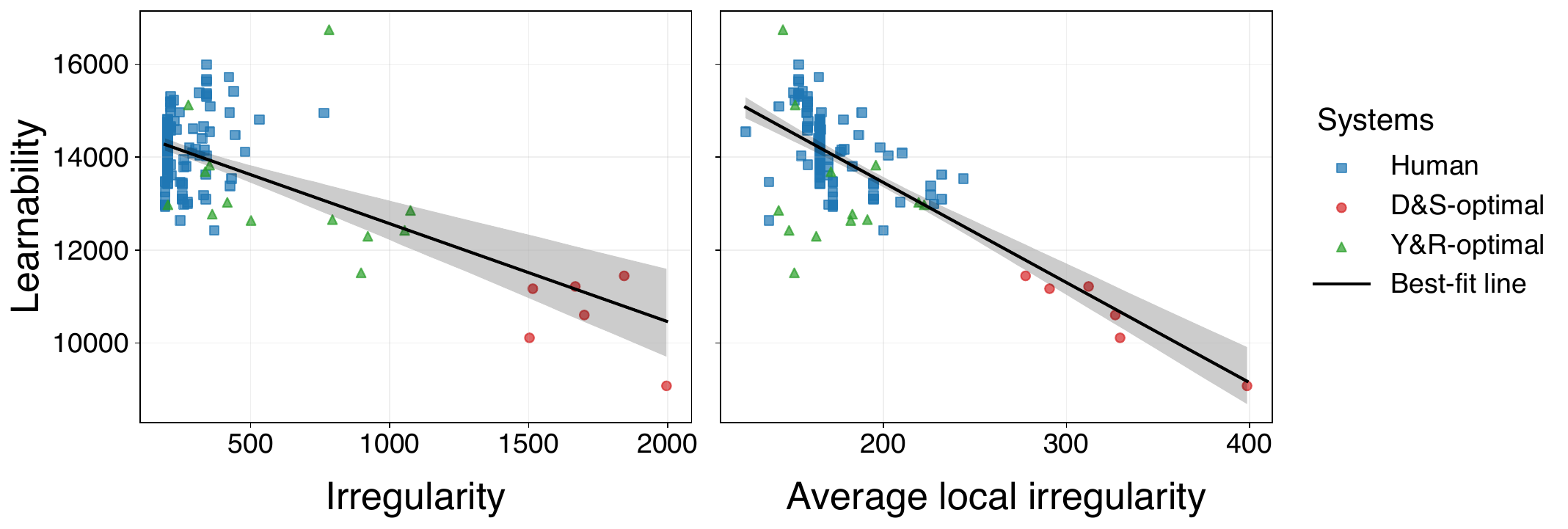}
    \caption{Irregularity (left) and average local irregularity (right)  against learnability for human (blue), D\&S optimal (red) and Y\&R optimal (green) systems, under the power law training distribution 
    and a uniform test distribution.
    }
    \label{fig:optimal_only}
\end{figure*}

Figure \ref{fig:optimal_only} (left) provides a closer look at human and highly regular languages. D\&S systems, despite being efficient in terms of lexicon size and length, have relatively high irregularity and hence low learnability, whereas Y\&R systems were constrained to be human-like and regular, and therefore have higher learnability.

There are however a few clear outliers, such as the highly learnable Y\&R system (the green triangle at the top of Figure \ref{fig:optimal_only}, left), which is more irregular than most human systems yet more learnable.
This could be because the current measure of regularity does not take into account the structure of the output space, but only the degree of overall regularity (i.e., how much form is reused across \textit{all} numerals).
However, it is plausible that learnability is largely affected by \textit{local} regularity (i.e., how much form is reused between consecutive numbers).
French, for instance, uses incosistent bases, but uses a consistent base for a given range of consecutive numbers: base 10 in the 30s up to the 60s (e.g., \textit{quar-ante cinq} $4*10+5$), base 20 in the 80s (e.g., \textit{quartre-vingt-un} $4*20+1$), and both in the 70s and 90s (e.g., \textit{quatre-vingt-dix-sept} $4*20+10+7$), and so on. 

We therefore compute \textit{average local regularity}, adapted from the previous regularity measure.
We first extract recursive numeral subsystems that correspond to 10  consecutive numerals from a given system (i.e., numerals in the ranges 1-10, 2-11, 3-12, ...).
We compute the irregularity for each of these (local) subsystems, then take the unweighted average to obtain the average local irregularity.
Figure \ref{fig:optimal_only} (right) shows that once local regularity is taken into account, the outliers appear more in line with the trend, suggesting that local regularity may indeed play key role in the learnability of a recursive numeral system for our RL agents.

\begin{figure}[t]
    \centering
    \includegraphics[width=0.47\textwidth]{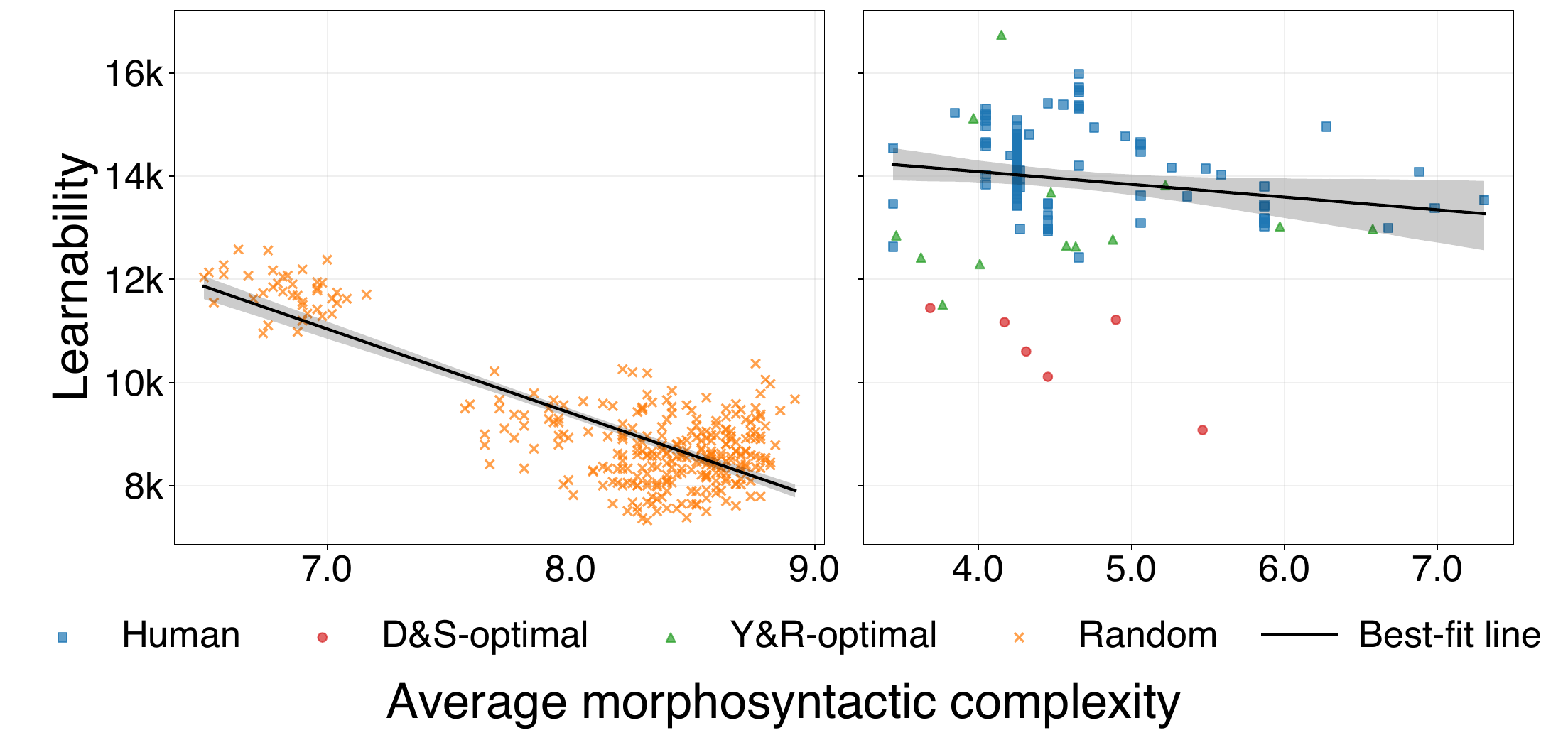}
    \vspace{-0.25em}
    \caption{Average morphosyntactic complexity and learnability for random baseline (left) and highly regular systems (right), under the power law training distribution and a uniform test distribution.}
    \label{fig:random_and_optimal_morpho}
\end{figure}

While it seems that regularity correlates well with learnability overall (Figure \ref{fig:optimal_random_flat_and_not}, right, $\beta = -0.828$) and for highly regular systems (Figure \ref{fig:optimal_only}, left, $\beta=-2.110$), random artificial systems do not follow the same trend (see the orange cluster in Figure \ref{fig:optimal_random_flat_and_not}, right), where the negative trend is weaker ($\beta=-0.259$ among this cluster).
For these highly irregular systems, we instead see an influence of average morphosyntactic complexity (i.e., average length): higher morphosyntactic complexity corresponds to lower learnability. 
This trend is almost absent for highly regular systems 
(Figure \ref{fig:random_and_optimal_morpho}).
This result suggests that in different areas of the space of possible recursive numeral systems, different measures can have a higher or lower impact on learnability. 
Specifically, the degree of regularity (i.e., the complexity of the rules involved) plays a large role when the system exhibits some regularity, whereas for highly irregular systems, it is so difficult to find any pattern that changes in regularity matter much less \citep[cf.][]{raviv2021}.
In this case, length becomes the main determinant for learnability since longer forms are harder to memorise.

In the next section, we test this idea more explicitly, and examine the regularity-learnability link in families of highly similar, human-like, regular recursive numeral systems with fixed average morphosyntactic complexity and lexicon size.

\section{Experiment 3: Learnability among highly regular, human-like systems}

\subsection{Experiment setup}
We adopt the same methods as the previous two experiments, but our samples now come from the most and least efficient systems in the \textit{local neighbourhoods} of human recursive numeral systems.
These are families of artificially generated systems in \citet{prasertsom2026recursivenumeralsystemshighly} that have the same digits, multipliers, arithmetic combinators ($+,-,*$) and numeral lengths as human recursive systems.
For example, Basque uses $10$ and $20$ as multipliers, and $91$ is expressed as $4*20+10+1$ (length = 7).
Thus, its alternative numerals of the same length are $\{10+4*20+1, 10+8*10+1, 20+7*10+1,
4*20+10+1, 7*10+20+1, 8*10+10+1\}$.
Basque's local neighbourhoods consist of systems comprised of combinations of these alternative numerals.
This effectively also means that systems in each local neighbourhood will all have the same lexicon size and average morphosyntactic complexity.
The samples only consist of the most and least regular systems obtained from a greedy algorithm.
Since some neighbourhoods feature very small variation, we only include 30 out of 37 neighbourhoods that have more than 10 systems.

\subsection{Results and discussion}
\begin{figure}[t]
    \centering
    \includegraphics[width=0.45\textwidth]{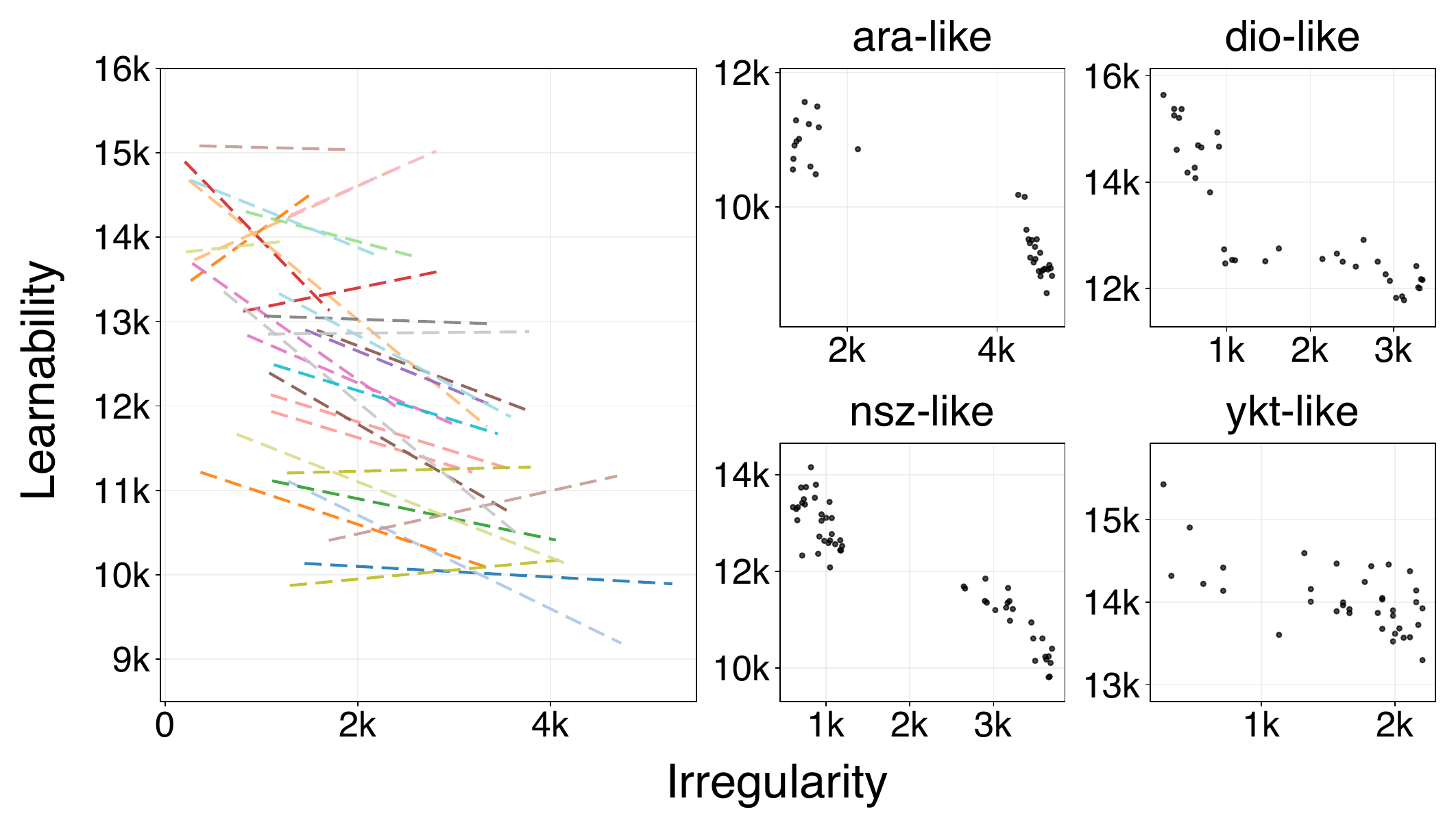}
    \vspace{-0.25em}
    \caption{Left: best-fit regression lines between irregularity and learnability for the higher-populated local neighbourhoods (one colour per neighbourhood) from \citet{prasertsom2026recursivenumeralsystemshighly}.
    Right: Irregularity and learnability of example local neighbourhoods (Arawak-like, Diola-like, Nahuatl-like and Yakut-like systems).}
    \label{fig:part3_full_plot}
\end{figure}

As predicted, we again found a negative correlation between irregularity and learnability in most of these neighbourhoods.
We fit a least-square regression predicting learnability from irregularity for each local neighbourhood, and found a negative trend in 21 out of 30 neighbourhoods (Figure \ref{fig:part3_full_plot}, left; Mean $\beta{}_\text{irregularity} = -0.249; \text{S.E.}=0.451$).
Figure \ref{fig:part3_full_plot} (right) gives illustrative examples from four neighbourhoods (see the full plot that includes every neighbourhood see \textcolor{blue!70!black}{\href{https://anonymous.4open.science/r/cogsci26_recnumsys-EF6B/README.md}{our anonymised GitHub repository}}.


\section{General discussion}
In this work, we explored whether learnability pressures can explain the cross-linguistic prevalence of regularity in recursive numeral systems.
In particular, using RL, we measured the learnability of attested and hypothetical recursive numeral systems with varying degrees of regularity, and tested whether regularity correlates with learnability.
We found that this is born out under the assumption that the relevant measure of learnability reflects the need to generalise from limited data to precisely express all integers. By contrast, it does not hold under the assumption that both the training and testing distributions are the power law communicative need distribution \citep{dehaene1992cross}.  
We additionally demonstrated that regularity has less influence on the learnability of highly irregular systems.
Our overall results thus corroborate the idea that learnability explains the cross-linguistic prevalence of regular recursive numeral systems.
They also point to the particular importance of generalisation in the learning of numeral systems, even if frequency of use is highly skewed.

We have shown the learnability advantage of regularity with RL. 
A natural prediction is that regularity should similarly confer a learnability advantage for humans in this domain.
However, in two artificial numeral learning experiments, \citet{holt2025learning} found little evidence of English-speaking adults' learning performance with different amounts of exposure to generalisable, regular rules.
While this may seem at odds with our prediction, it is worth noting that the numerals used in this study were in the lower ranges especially during training (1--10 at train, 11-20 at test), and there was only one round of learning and generalisation.
In other words, unlike humans in real-life situations who have a real need to generalise to large numerals, the participants were not presented with any direct pressure to perform well at generalisation.
Future designs might implement this pressures artificially, e.g., by iteratively alternating between learning and generalisation.

Another issue worth exploring concerns the role of input orders.
Our model treats the input as random samples from the distribution of use. 
However, recursive numerals are typically learned through counting routines  \citep{fuson2012children,carey2000origin,carey2019ontogenetic}, perhaps even necessarily so \citep{pitt2022exact}.
One way counting routines could facilitate learning is by artificially boosting the frequency of numerals over 10 (e.g. in the range 10--30), which are not frequently used for communication according to the power law, but are often part of counting routines.
This would expose learners to a range of numerals from which highly generalisable rules can be extracted.
For instance, in Mandarin, numerals in the range of 1--10 will not allow the learner to extract any general pattern because they all have dedicated forms, whereas the range 1--30 contains all the patterns required for all numerals up to 99 (e.g., you can reuse the rules generating $3*10+1$ for any numeral of the same structure like $4*10+5$).\footnote{This is also reflected in the local regularity of Mandarin numerals in these ranges.
It increases as more numbers  are covered but no longer increase after 30 ($56.219$ for 1-10, $201.192$ for 1-20, $241.039$ for 1-30, 1-40, 1-50 up to 1-99).}
It would be possible to investigate this with the current model, or a Markov Decision Process \citep{Sutton1998} where an agent is exposed to consecutive numbers in a single episode.

Our results also suggest that the relationship between regularity and learnability is likely non-linear, and other aspects of complexity (like length) might influence learnability more in some parts of the space.
Going beyond regularity, an interesting future direction would be to identify the influence on learnability not only of morphosyntactic complexity/length, but also other pressures such as processing complexity, hypothesised by \citet{prasertsom2026recursivenumeralsystemshighly} to also trade off with regularity among highly efficient numeral systems.
The fact that learnability is multiply determined also means that a learnability-based measure of complexity \citep[e.g.,][]{osmelak2026systematicityformsmeaningslanguages} may be of limited use as a substitute for a more transparent measure of a specific aspect of complexity (like our regularity measures), as learnability may depend on different features of systems in different regions of the space.


Finally and more broadly, our study may have implications for the evolution of regularity in general.
We find that regularity is most useful when the learning metric requires generalisation beyond the learning distribution (i.e., when the train and test distributions are different).
This shows that regular systems in general enjoy the benefits of being robust to different need distributions. 
We suggest that, besides the learning vs. generalising distribution distinction, this distinction could also be framed in terms of \textit{changing} communicative needs.
In addition to being robust to limited learning samples (cf. the \textit{transmission bottleneck}; \citealp{smith2003iterated}), then, regular systems may also be selected in language evolution because they are a robust solution to changes in communicative needs over individual and historical times \citep{karjus2020quantifying}.

\section{Acknowledgments}

P.P. is supported by Anandamahidol Foundation, Thailand. 
A.S. is supported by the Swedish Research Council. The computations were enabled by resources provided by Chalmers e-Commons at Chalmers.


\printbibliography

@article{moreton2017phonological,
  title={Phonological concept learning},
  author={Moreton, Elliott and Pater, Joe and Pertsova, Katya},
  journal={Cognitive science},
  volume={41},
  number={1},
  pages={4--69},
  year={2017},
  publisher={Wiley Online Library}
}

@article{wang2025learning,
  title={The learning bias for cross-category harmony is sensitive to semantic similarity: Evidence from artificial language learning experiments},
  author={Wang, Fang and Kirby, Simon and Culbertson, Jennifer},
  journal={Language},
  volume={101},
  number={1},
  pages={109--150},
  year={2025},
  publisher={Linguistic Society of America}
}

@article{smith2003iterated,
  title={Iterated learning: A framework for the emergence of language},
  author={Smith, Kenny and Kirby, Simon and Brighton, Henry},
  journal={Artificial life},
  volume={9},
  number={4},
  pages={371--386},
  year={2003},
  publisher={MIT Press}
}

@article{smith2010eliminating,
  title={Eliminating unpredictable variation through iterated learning},
  author={Smith, Kenny and Wonnacott, Elizabeth},
  journal={Cognition},
  volume={116},
  number={3},
  pages={444--449},
  year={2010},
  publisher={Elsevier}
}

@article{keogh2024predictability,
  title={Predictability and variation in language are differentially affected by learning and production},
  author={Keogh, Aislinn and Kirby, Simon and Culbertson, Jennifer},
  journal={Cognitive Science},
  volume={48},
  number={4},
  pages={e13435},
  year={2024},
  publisher={Wiley Online Library}
}

@misc{osmelak2026systematicityformsmeaningslanguages,
      title={Systematicity between Forms and Meanings across Languages Supports Efficient Communication}, 
      author={Doreen Osmelak and Yang Xu and Michael Hahn and Kate McCurdy},
      year={2026},
      eprint={2601.17181},
      archivePrefix={arXiv},
      primaryClass={cs.CL},
      url={https://arxiv.org/abs/2601.17181}, 
}

@book{croft2002typology,
  title={Typology and universals},
  author={Croft, William},
  year={2002},
  publisher={Cambridge university press}
}

@article{evans2009myth,
  title={The myth of language universals: Language diversity and its importance for cognitive science},
  author={Evans, Nicholas and Levinson, Stephen C},
  journal={Behavioral and brain sciences},
  volume={32},
  number={5},
  pages={429--448},
  year={2009},
  publisher={Cambridge University Press}
}

@article{karjus2020quantifying,
  title={Quantifying the dynamics of topical fluctuations in language},
  author={Karjus, Andres and Blythe, Richard A and Kirby, Simon and Smith, Kenny},
  journal={Language Dynamics and Change},
  volume={10},
  number={1},
  pages={86--125},
  year={2020},
  publisher={Brill}
}

@article{pitt2022exact,
  title={Exact number concepts are limited to the verbal count range},
  author={Pitt, Benjamin and Gibson, Edward and Piantadosi, Steven T},
  journal={Psychological Science},
  volume={33},
  number={3},
  pages={371--381},
  year={2022},
  publisher={Sage Publications Sage CA: Los Angeles, CA}
}

@article{carey2000origin,
  title={The origin of concepts},
  author={Carey, Susan},
  journal={Journal of Cognition and Development},
  volume={1},
  number={1},
  pages={37--41},
  year={2000},
  publisher={Taylor \& Francis}
}

@article{carey2019ontogenetic,
  title={Ontogenetic origins of human integer representations},
  author={Carey, Susan and Barner, David},
  journal={Trends in cognitive sciences},
  volume={23},
  number={10},
  pages={823--835},
  year={2019},
  publisher={Elsevier}
}

@book{fuson2012children,
  title={Children’s counting and concepts of number},
  author={Fuson, Karen C},
  year={2012},
  publisher={Springer Science \& Business Media}
}

@article{denic2024recursive,
  title={Recursive Numeral Systems Optimize the Trade-off Between Lexicon Size and Average Morphosyntactic Complexity},
  author={Deni{\'c}, Milica and Szymanik, Jakub},
  journal={Cognitive Science},
  volume={48},
  number={3},
  pages={e13424},
  year={2024},
  publisher={Wiley Online Library},
doi={https://doi.org/10.1111/cogs.13424}
}

@article{kemp2018semantic,
  title={Semantic typology and efficient communication},
url={https://doi.org/10.1146/annurev-linguistics-011817-045406},
  author={Kemp, Charles and Xu, Yang and Regier, Terry},
  journal={Annual Review of Linguistics},
  volume={4},
  number={1},
  pages={109--128},
  year={2018},
  publisher={Annual Reviews}
}

@inproceedings{yang2025re,
  title={Re-examining the tradeoff between lexicon size and average morphosyntactic complexity in recursive numeral systems},
  author={Yang, David and Regier, Terry},
  booktitle={Proceedings of the Annual Meeting of the Cognitive Science Society},
  volume={47},
  year={2025},
url={https://escholarship.org/uc/item/5k8646v4}
}

@article{holt2025learning,
  title={Learning a Novel Number System: The Role of Compositional Rules and Counting Procedures},
  author={Holt, Sebastian and Barner, David},
  journal={Cognitive Science},
  volume={49},
  number={6},
  pages={e70071},
  year={2025},
url={https://doi.org/10.1111/cogs.70071},
  publisher={Wiley Online Library}
}

@article{xu2020numeral,
  title={Numeral systems across languages support efficient communication: From approximate numerosity to recursion},
  author={Xu, Yang and Liu, Emmy and Regier, Terry},
  journal={Open Mind},
  volume={4},
  pages={57--70},
  year={2020},
url={https://doi.org/10.1162/opmi_a_00034}
}

@phdthesis{brighton2003simplicity,
  title        = {Simplicity as a Driving Force in Linguistic Evolution},
  author       = {Brighton, Henry},
  year         = {2003},
  school       = {The University of Edinburgh},
  type         = {PhD thesis},
  url          = {http://hdl.handle.net/1842/23810}
}

@article{dehaene1992cross,
  title={Cross-linguistic regularities in the frequency of number words},
  author={Dehaene, Stanislas and Mehler, Jacques},
  journal={Cognition},
  volume={43},
  number={1},
  pages={1--29},
  year={1992},
  publisher={Elsevier},
url={https://doi.org/10.1016/0010-0277(92)90030-L}
}

@inproceedings{carlsson2021learning,
  title={Learning approximate and exact numeral systems via reinforcement learning},
  author={Carlsson, Emil and Dubhashi, Devdatt and Johansson, Fredrik D},
  booktitle={Proceedings of the Annual Meeting of the Cognitive Science Society},
 volume ={43},
  year={2021}
}

@book{mihovFiniteStateTechniquesAutomata2019,
  title = {Finite-{{State Techniques}}: {{Automata}}, {{Transducers}} and {{Bimachines}}},
  shorttitle = {Finite-{{State Techniques}}},
  author = {Mihov, Stoyan and Schulz, Klaus U.},
  date = {2019-08-01},
year={2019},
  edition = {1},
  publisher = {Cambridge University Press},
  doi = {10.1017/9781108756945},
  urldate = {2025-09-24},
  isbn = {978-1-108-75694-5 978-1-108-48541-8}
}

@article{culbertson2012learning,
  title={Learning biases predict a word order universal},
  author={Culbertson, Jennifer and Smolensky, Paul and Legendre, G{\'e}raldine},
  journal={Cognition},
  volume={122},
  number={3},
  pages={306--329},
  year={2012},
  publisher={Elsevier}
}

@article{steinert2020ease,
  title={Ease of learning explains semantic universals},
  author={Steinert-Threlkeld, Shane and Szymanik, Jakub},
  journal={Cognition},
  volume={195},
  pages={104076},
  year={2020},
  publisher={Elsevier},
  doi={10.1016/j.cognition.2019.104076},
  url={https://doi.org/10.1016/j.cognition.2019.104076}
}

@article{steinert2019learnability,
  title={Learnability and semantic universals},
  author={Steinert-Threlkeld, Shane and Szymanik, Jakub and others},
  journal={Semantics and Pragmatics},
  volume={12},
  pages={401--435},
  year={2019},
  doi={10.3765/sp.12.4},
  url={https://dx.doi.org/10.3765/sp.12.4}
}

@article{shepard1961learning,
  title={Learning and memorization of classifications.},
  author={Shepard, Roger N and Hovland, Carl I and Jenkins, Herbert M},
  journal={Psychological monographs: General and applied},
  volume={75},
  number={13},
  pages={1},
  year={1961},
  publisher={American Psychological Association}
}

@misc{prasertsom2026recursivenumeralsystemshighly,
      title={Recursive numeral systems are highly regular and easy to process}, 
      author={Ponrawee Prasertsom and Andrea Silvi and Jennifer Culbertson and Moa Johansson and Devdatt Dubhashi and Kenny Smith},
      year={2026},
      note   = {arXiv:2501.01234. To appear in Proceedings of the EACL},
      eprint={2510.27049},
      archivePrefix={arXiv},
      primaryClass={cs.CL},
      url={https://arxiv.org/abs/2510.27049}, 
}

@article{gibson_how_2019,
    title = {How {Efficiency} {Shapes} {Human} {Language}},
    volume = {23},
    issn = {13646613},
    url = {https://linkinghub.elsevier.com/retrieve/pii/S1364661319300580},
    doi = {10.1016/j.tics.2019.02.003},
    language = {en},
    number = {5},
    urldate = {2023-06-12},
    journal = {Trends in Cognitive Sciences},
    author = {Gibson, Edward and Futrell, Richard and Piantadosi, Steven P. and Dautriche, Isabelle and Mahowald, Kyle and Bergen, Leon and Levy, Roger},
    month = may,
    year = {2019},
    pages = {389--407},
}

@article{frank2008piraha,
title = {Number as a cognitive technology: Evidence from Pirahã language and cognition},
journal = {Cognition},
volume = {108},
number = {3},
pages = {819-824},
year = {2008},
issn = {0010-0277},
doi = {https://doi.org/10.1016/j.cognition.2008.04.007},
url = {https://www.sciencedirect.com/science/article/pii/S0010027708001042},
author = {Michael C. Frank and Daniel L. Everett and Evelina Fedorenko and Edward Gibson}
}

@inproceedings{foerster2016,
author = {Foerster, Jakob N. and Assael, Yannis M. and de Freitas, Nando and Whiteson, Shimon},
title = {Learning to communicate with Deep multi-agent reinforcement learning},
year = {2016},
isbn = {9781510838819},
publisher = {Curran Associates Inc.},
address = {Red Hook, NY, USA},
booktitle = {Proceedings of the 30th International Conference on Neural Information Processing Systems},
pages = {2145–2153},
numpages = {9},
location = {Barcelona, Spain},
series = {NIPS'16}
}

@inproceedings{LazaridouPB17,
  author       = {Angeliki Lazaridou and
                  Alexander Peysakhovich and
                  Marco Baroni},
  title        = {Multi-Agent Cooperation and the Emergence of (Natural) Language},
  booktitle    = {5th International Conference on Learning Representations, {ICLR} 2017,
                  Toulon, France, April 24-26, 2017, Conference Track Proceedings},
  publisher    = {OpenReview.net},
  year         = {2017},
  url          = {https://openreview.net/forum?id=Hk8N3Sclg},
  bibsource    = {dblp computer science bibliography, https://dblp.org}
}

@inproceedings{mordatch2018,
author = {Mordatch, Igor and Abbeel, Pieter},
title = {Emergence of grounded compositional language in multi-agent populations},
year = {2018},
isbn = {978-1-57735-800-8},
publisher = {AAAI Press},
booktitle = {Proceedings of the Thirty-Second AAAI Conference on Artificial Intelligence and Thirtieth Innovative Applications of Artificial Intelligence Conference and Eighth AAAI Symposium on Educational Advances in Artificial Intelligence},
articleno = {183},
numpages = {8},
location = {New Orleans, Louisiana, USA},
series = {AAAI'18/IAAI'18/EAAI'18}
}

@article{carlsson2024,
    author = {Carlsson, Emil and Dubhashi, Devdatt and Regier, Terry},
    title = {Cultural evolution via iterated learning and communication explains efficient color naming systems},
    journal = {Journal of Language Evolution},
    volume = {9},
    number = {1-2},
    pages = {49-66},
    year = {2024},
    month = {11},
    issn = {2058-458X},
    doi = {10.1093/jole/lzae010},
    url = {https://doi.org/10.1093/jole/lzae010},
    eprint = {https://academic.oup.com/jole/article-pdf/9/1-2/49/60792015/lzae010.pdf},
}

@article{taylor2009,
author = {Taylor, Matthew E. and Stone, Peter},
title = {Transfer Learning for Reinforcement Learning Domains: A Survey},
year = {2009},
issue_date = {12/1/2009},
publisher = {JMLR.org},
volume = {10},
issn = {1532-4435},
journal = {J. Mach. Learn. Res.},
month = dec,
pages = {1633–1685},
numpages = {53}
}

@inproceedings{agarwal2021,
author = {Agarwal, Rishabh and Schwarzer, Max and Castro, Pablo Samuel and Courville, Aaron and Bellemare, Marc G.},
title = {Deep reinforcement learning at the edge of the statistical precipice},
year = {2021},
isbn = {9781713845393},
publisher = {Curran Associates Inc.},
address = {Red Hook, NY, USA},
booktitle = {Proceedings of the 35th International Conference on Neural Information Processing Systems},
articleno = {2244},
numpages = {17},
series = {NIPS '21}
}

@book{Sutton1998,
  author = {Sutton, Richard S. and Barto, Andrew G.},
  edition = {Second},
  publisher = {The MIT Press},
  title = {Reinforcement Learning: An Introduction}, 
  year = {1998}
  }

@article{agrawal1995continuum,
author = {Agrawal, Rajeev},
title = {The Continuum-Armed Bandit Problem},
year = {1995},
issue_date = {Nov. 1995},
publisher = {Society for Industrial and Applied Mathematics},
address = {USA},
volume = {33},
number = {6},
issn = {0363-0129},
url = {https://doi.org/10.1137/S0363012992237273},
doi = {10.1137/S0363012992237273},
journal = {SIAM J. Control Optim.},
month = nov,
pages = {1926–1951},
numpages = {26}
}

@book{lattimore2020bandit,
  title={Bandit algorithms},
  author={Lattimore, Tor and Szepesv{\'a}ri, Csaba},
  year={2020},
  publisher={Cambridge University Press}
}

@article{hochreiter1997long,
  title={Long Short-term Memory},
  author={Hochreiter, S},
  journal={Neural Computation MIT-Press},
  year={1997}
}

@article{williams1992reinforce,
author = {Williams, Ronald J.},
title = {Simple Statistical Gradient-Following Algorithms for Connectionist Reinforcement Learning},
year = {1992},
issue_date = {May 1992},
publisher = {Kluwer Academic Publishers},
address = {USA},
volume = {8},
number = {3–4},
issn = {0885-6125},
url = {https://doi.org/10.1007/BF00992696},
doi = {10.1007/BF00992696},
journal = {Mach. Learn.},
month = may,
pages = {229–256},
numpages = {28}
}

@article{guo2019emergence,
  author       = {Shangmin Guo and
                  Yi Ren and
                  Serhii Havrylov and
                  Stella Frank and
                  Ivan Titov and
                  Kenny Smith},
  title        = {The Emergence of Compositional Languages for Numeric Concepts Through
                  Iterated Learning in Neural Agents},
  journal      = {CoRR},
  volume       = {abs/1910.05291},
  year         = {2019},
  url          = {http://arxiv.org/abs/1910.05291},
  eprinttype    = {arXiv},
  eprint       = {1910.05291},
  bibsource    = {dblp computer science bibliography, https://dblp.org}
}

@article{kageback2020eff,
    doi = {10.1371/journal.pone.0234894},
    author = {Kågebäck, Mikael AND Carlsson, Emil AND Dubhashi, Devdatt AND Sayeed, Asad},
    journal = {PLOS ONE},
    publisher = {Public Library of Science},
    title = {A reinforcement-learning approach to efficient communication},
    year = {2020},
    month = {07},
    volume = {15},
    url = {https://doi.org/10.1371/journal.pone.0234894},
    pages = {1-26},
    number = {7},

}

@phdthesis{kakade2003sample,
  title={On the sample complexity of reinforcement learning},
  author={Kakade, Sham M},
  year={2003},
  school={University College London},
    address      = {London, United Kingdom},
  type         = {Doctoral dissertation}
}

@article{raviv2021,
title = {What makes a language easy to learn? A preregistered study on how systematic structure and community size affect language learnability},
journal = {Cognition},
volume = {210},
pages = {104620},
year = {2021},
issn = {0010-0277},
doi = {https://doi.org/10.1016/j.cognition.2021.104620},
url = {https://www.sciencedirect.com/science/article/pii/S0010027721000391},
author = {Limor Raviv and Marianne {de Heer Kloots} and Antje Meyer}
}

@article{Galke_2024,
   title={Deep neural networks and humans both benefit from compositional language structure},
   volume={15},
   ISSN={2041-1723},
   url={http://dx.doi.org/10.1038/s41467-024-55158-1},
   DOI={10.1038/s41467-024-55158-1},
   number={1},
   journal={Nature Communications},
   publisher={Springer Science and Business Media LLC},
   author={Galke, Lukas and Ram, Yoav and Raviv, Limor},
   year={2024},
   month=dec }

@inproceedings{silvi2025le,
  title={Learning Efficient Recursive Numeral Systems via Reinforcement Learning},
  author={Silvi, Andrea and Thomas, Jonathan and Carlsson, Emil and Dubhashi, Devdatt and Johansson, Moa},
  booktitle={Proceedings of the Annual Meeting of the Cognitive Science Society},
  volume={47},
  year={2025},
url={https://escholarship.org/uc/item/3cc5053z}
}

@article{CHATER200319,
title = {Simplicity: a unifying principle in cognitive science?},
journal = {Trends in Cognitive Sciences},
volume = {7},
number = {1},
pages = {19-22},
year = {2003},
issn = {1364-6613},
doi = {https://doi.org/10.1016/S1364-6613(02)00005-0},
url = {https://www.sciencedirect.com/science/article/pii/S1364661302000050},
author = {Nick Chater and Paul Vitányi}
}

\end{document}